\documentclass[journal]{IEEEtran}

\usepackage{amsmath,amsfonts}
\usepackage{algorithm}
\usepackage{algpseudocode}
\usepackage{array}
\usepackage[]{subfig}
\usepackage{textcomp}
\usepackage{stfloats}
\usepackage{url}
\usepackage{verbatim}

\usepackage{graphicx}
\usepackage{cite}
\usepackage{hyperref}
\usepackage{bm}
\usepackage{makecell}
\usepackage{footnote}
\usepackage{float}
\usepackage{subfig}
\usepackage{bbding}
\usepackage{ragged2e}
\usepackage{amsthm}
\usepackage{dsfont}
\usepackage{lipsum} 
\usepackage{booktabs}
\usepackage{multirow}
\usepackage{xcolor}
\usepackage{makecell}
\usepackage[normalem]{ulem}
\useunder{\uline}{\ul}{}

\usepackage{enumitem} 
\usepackage{pifont}

\makeatletter

\newcommand{\Rmnum}[1]{\expandafter\@slowromancap\romannumeral #1@}
\makeatother

\theoremstyle{definition}

\usepackage{cuted}

\usepackage[switch]{lineno}

\begin{document}

\title{Region-Aware CAM: High-Resolution Weakly-Supervised Defect Segmentation via Salient Region Perception}

\author{Hang-Cheng~Dong, Lu Zou, Bingguo~Liu, Dong~Ye, Guodong Liu


\thanks{\textit{(Corresponding authors: Bingguo Liu)}}
\thanks{Hang-Cheng~Dong, Lu Zou, Bingguo~Liu, Dong~Ye and Guodong Liu are with School of Instrumentation Science and Engineering, Harbin Institute of Technology, Harbin 150001, China (email: liu\_bingguo@hit.edu.cn).}
\thanks{Hang-Cheng~Dong, Lu Zou, Bingguo~Liu and Guodong Liu are with Harbin Institute of Technology Suzhou Research Institute, Suzhou 215000, China.}

}


\maketitle

\begin{abstract}

Surface defect detection plays a critical role in industrial quality inspection. Recent advances in artificial intelligence have significantly enhanced the automation level of detection processes. However, conventional semantic segmentation and object detection models heavily rely on large-scale annotated datasets, which conflicts with the practical requirements of defect detection tasks. This paper proposes a novel weakly supervised semantic segmentation framework comprising two key components: a region-aware class activation map (CAM) and pseudo-label training. To address the limitations of existing CAM methods, especially low-resolution thermal maps, and insufficient detail preservation, we introduce filtering-guided backpropagation (FGBP), which refines target regions by filtering gradient magnitudes to identify areas with higher relevance to defects. Building upon this, we further develop a region-aware weighted module to enhance spatial precision. Finally, pseudo-label segmentation is implemented to refine the model's performance iteratively. Comprehensive experiments on industrial defect datasets demonstrate the superiority of our method. The proposed framework effectively bridges the gap between weakly supervised learning and high-precision defect segmentation, offering a practical solution for resource-constrained industrial scenarios.

\end{abstract}


\section{Introduction}
 \IEEEPARstart{S}{urface} defect detection plays a critical role in intelligent manufacturing systems~\cite{gao2022review}, serving as a vital component for quality control. Conventional approaches based on machine vision and image processing techniques rely heavily on handcrafted feature extractors~\cite{GOLNABI2007630}, which struggle to meet the requirements for automated defect detection in dynamic environments with complex backgrounds. The emergence of deep learning methodologies has shed light on intelligent defect detection systems by enabling automated feature learning.

Unfortunately, the requirement of deep learning techniques for labeled samples hinders their large-scale application in the field of surface defect detection~\cite{ren2022state}. In the case of defects in industrial products, the occurrence of defects is rare because they are designed to avoid such situations. Meanwhile, the long accumulation period required for specific defect categories poses a significant challenge for dataset construction~\cite{ksdd2}. This weakness in dataset construction can also affect the performance of deep learning algorithms. In particular, the scarcity of certain categories can lead to a long-tail~\cite{longtail} distribution problem, which induces model bias and affects model performance.

\begin{figure}[t]
    \centering
    \includegraphics[scale=0.2]{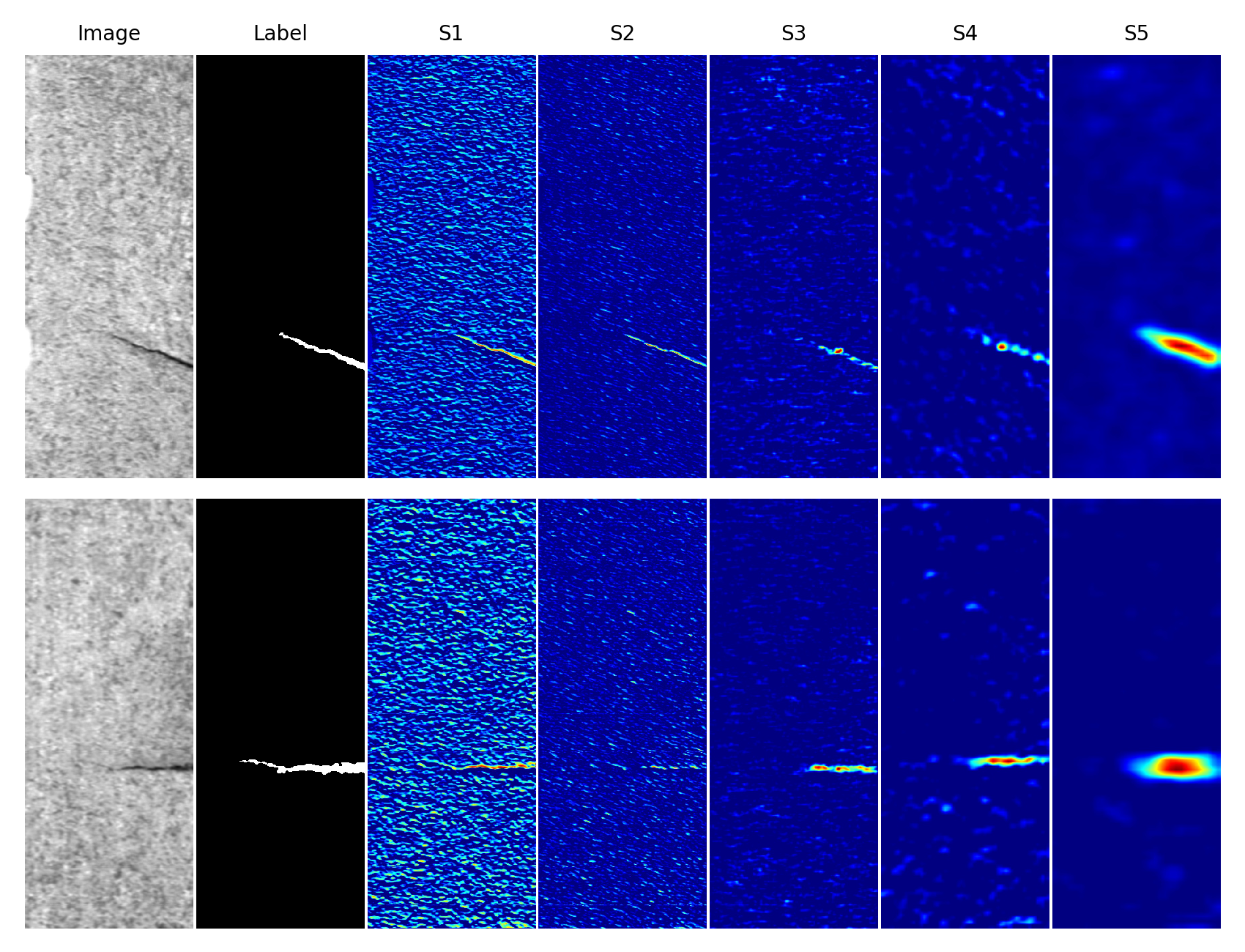}
    \caption{Heatmaps generated by Grad-CAM across 5 VGG16 convolutional stages, based on KSDD dataset.}
    \label{fig1:gradcam_KSDD}
\end{figure}

Furthermore, conventional fully-supervised learning paradigms impose stringent requirements on annotation quality, especially for critical tasks like object detection~\cite{od2022review} and semantic segmentation~\cite{sultana2020evolution}. In defect detection applications, semantic segmentation demands pixel-level annotations that necessitate specialized domain expertise~\cite{liu2023survey}. This creates a paradoxical situation where technically skilled annotators often lack the required materials science knowledge, while domain experts may be unfamiliar with annotation tools. Compounding these challenges, ambiguous defect boundaries frequently introduce label noise. The combined effect of scarce defect samples and labor-intensive annotation processes significantly hinders progress in intelligent quality inspection systems, highlighting the urgent need for approaches that reduce dependence on pixel-level supervision.

Recent advances in unsupervised and weakly supervised learning offer promising alternatives for surface defect detection~\cite{tao2022deep}. While unsupervised methods eliminate annotation requirements by utilizing readily available normal samples, their practical application is limited by suboptimal performance and weak generalization capabilities. In contrast, weakly-supervised approaches demonstrate superior effectiveness, particularly for semantic segmentation tasks in defect detection. This study focuses on advancing weakly-supervised semantic segmentation (WSSS) techniques for enhanced defect detection performance.

Weakly-supervised semantic segmentation primarily employs Class Activation Mapping (CAM) techniques~\cite{CAM}, which generate class-specific heatmaps by weighting feature maps from the final convolutional layer. This approach enables target region localization using only image-level labels. Recent developments include the application of CAM with HRNet for defect detection~\cite{CADN}, achieving improved heatmap resolution through high-resolution network architectures. The Grad-CAM~\cite{GradCAM} introduced gradient-based channel weighting, extending CAM compatibility to arbitrary CNN architectures. While deeper layers typically exhibit reduced spatial resolution but cleaner background separation due to pooling operations, LayerCAM~\cite{layercam} addresses shallow layer noise suppression through gradient vector weighting. However, this method introduces fragmented features due to inconsistency between the feature map and gradient responses, presenting new challenges for precise segmentation.

To address the challenge of obtaining high-resolution activation maps from CNN feature representations, we propose a high-resolution weakly supervised semantic segmentation method for surface defect detection. As illustrated in Fig.~\ref{fig1:gradcam_KSDD}, shallow CNN layers inherently preserve higher spatial resolution but suffer from substantial non-target noise contamination. While LayerCAM attempts to suppress background interference by multiplying gradient signs with feature maps, this approach neglects the quantitative influence of gradient magnitudes. Our analysis reveals that the magnitude of gradients associated with features strongly correlates with their probability of belonging to target regions. Namely, the higher the value of the gradient, the higher the likelihood that the defect will be localized.

Therefore, we first propose the filtering-guide backpropagation (FGBP) method based on filtering the background noise of non-interested regions, and this technique can be used to improve the performance of various types of gradient-based heat maps. Furthermore, we propose region-aware class activation maps (RA-CAM) that use the proposed filtering to guide backpropagation to remove the interference of non-target regions on the weights of the feature maps, which greatly improves the segmentation resolution of the target regions. Finally, we train the segmentation model using the obtained pseudo labels and find that pseudo-label training is also important for further improving the defect detection performance.

In summary, our contributions are threefold:

{\small $\bullet$} {We propose a weakly supervised semantic segmentation method for automated industrial surface defect segmentation, consisting of heatmap generation and pseudo-label training steps, requiring only image-level annotations.}

{\small $\bullet$} We propose the filtering-guide backpropagation (FGBP) method, which filters out interference from non-target regions (background noise) in backpropagation and can be used as a plug-in for porting to other methods with similar processes.

{\small $\bullet$}We develope a class activation map based on salient region perception, termed RA-CAM. By eliminating interference from non-target regions in feature map weighting, high-resolution segmentation of defect regions is achieved.

\section{Related Works}

\subsection{Class Activation Maps}

CAM~\cite{CAM} were originally proposed as an interpretable method for deep learning to generate semantic interpretations of convolutional neural networks~\cite{fan}. Since CAM methods can provide rough pixel labels, they are often used in weakly supervised semantic segmentation or as starting seeds for pseudo-labeling. The methodology operates by weighting feature maps from the final convolutional layer, followed by channel-wise aggregation and upsampling to produce saliency maps. These maps approximate target regions by highlighting areas with high activation intensities. While Grad-CAM~\cite{GradCAM} pioneered gradient-based channel importance estimation for architectural flexibility. Subsequent innovations have developed enhanced weighting strategies through distinct theoretical frameworks. Grad-CAM++~\cite{Grad-cam++} theoretically grounds its formulation in the premise that positive gradient magnitudes correlate strongly with target-specific feature relevance. Moreover, XGrad-CAM~\cite{Axiom-based} introduces axiomatic mathematical principles to rectify weight calculation in ReLU-activated networks, ensuring gradient consistency during feature aggregation. Beyond gradient analysis, Lift-CAM~\cite{liftcam} and Relevance-CAM~\cite{RelevanceCAM} employ layer-wise relevance propagation~\cite{LRP} to quantify channel contributions. Alternative paradigms bypass gradient computation by directly measuring feature impact on model outputs. Score-CAM~\cite{scorecam} quantifies channel significance through activation masking experiments, while Ablation-CAM~\cite{ablationcam} systematically evaluates performance degradation during feature suppression. In addition, GroupCAM~\cite{Group-CAM} and FSG-CAM~\cite{FSGCAM} is based on similar principles. The Fullgrad~\cite{fullgrad} method analyzes the information in shallow feature maps and proposes to fuse the gradients of different layers to obtain heat maps. LayerCAM~\cite{layercam} does this by employing gradient matrices for weighting instead of the channel-by-channel linear combinations of GradCAM. NFF-CAM~\cite{zhou2025non} explores the effect of multi-scale input features, showing that adjusting the scale of the inputs helps to generate higher resolution heatmaps. SESS~\cite{tursun2022sess}, on the other hand, splits the original input into several small pieces and finally puts the heatmaps together, which is used to obtain more detailed heatmaps. However, these methods do not take into account the fact that the generated weights are always disturbed by non-target regions or background noise, which results in weights that can be distorted to some extent.

\subsection{Deep Learning Based Defect Inspection}


Compared to traditional image processing-based defect detection methods, deep learning offers greater potential for handling complex environments and backgrounds. In particular, object detection~\cite{yolov11,ren2016faster} and semantic segmentation~\cite{UNet,deeplabv3+} models are the most widely adopted deep learning frameworks. In object detection, \cite{kaikouxiao} proposed a three-stage method using an improved Faster R-CNN to locate fastener defects in high-speed railways. \cite{cui2021sddnet} introduced SDDNet for steel surface defects, incorporating a feature retention block (FRB) and skip dense connection module (SDCM) to address texture variations and small defects. \cite{ma2024hierarchical} developed a hierarchical attention mechanism for bearing surface defects, weighting features across texture, semantic, and instance levels. For semantic segmentation, \cite{dong2019pga119} proposed PGANet with a pyramid feature fusion module and global context attention to propagate multi-level defect features. \cite{yang2021automatic115} improved UNet with feature fusion and attention modules for welding defect detection. \cite{yang2021automatic115} designed STDC-Net, a lightweight encoder-decoder model using densely weighted connections and auxiliary boundary supervision to preserve edge details.

Classification-based approaches require simpler annotations. Based on the classification model, weakly supervised segmentation methods can perform segmentation or localization tasks with only image-level labels. \cite{LIU2021102008fabric} added a designed null convolutional spatial attention mechanism to the classification model to obtain higher quality thermograms to segment fabric surface defects. \cite{CADN} adopted HRNet as the backbone and used model distillation to obtain models with smaller parameter counts to adapt to the lighter-weight requirements of defect detection tasks. For the student model, a high-resolution feature layer was designed to get a thermogram that retains more details. Liu et al.~\cite{liu2022explainable} designed a multi-scale feature fusion module for laser welding defect detection and used CAM to demonstrate the interpretability of the model.

\section{Method}

\begin{figure}[ht]
    \centering
    \includegraphics[width=1\linewidth]{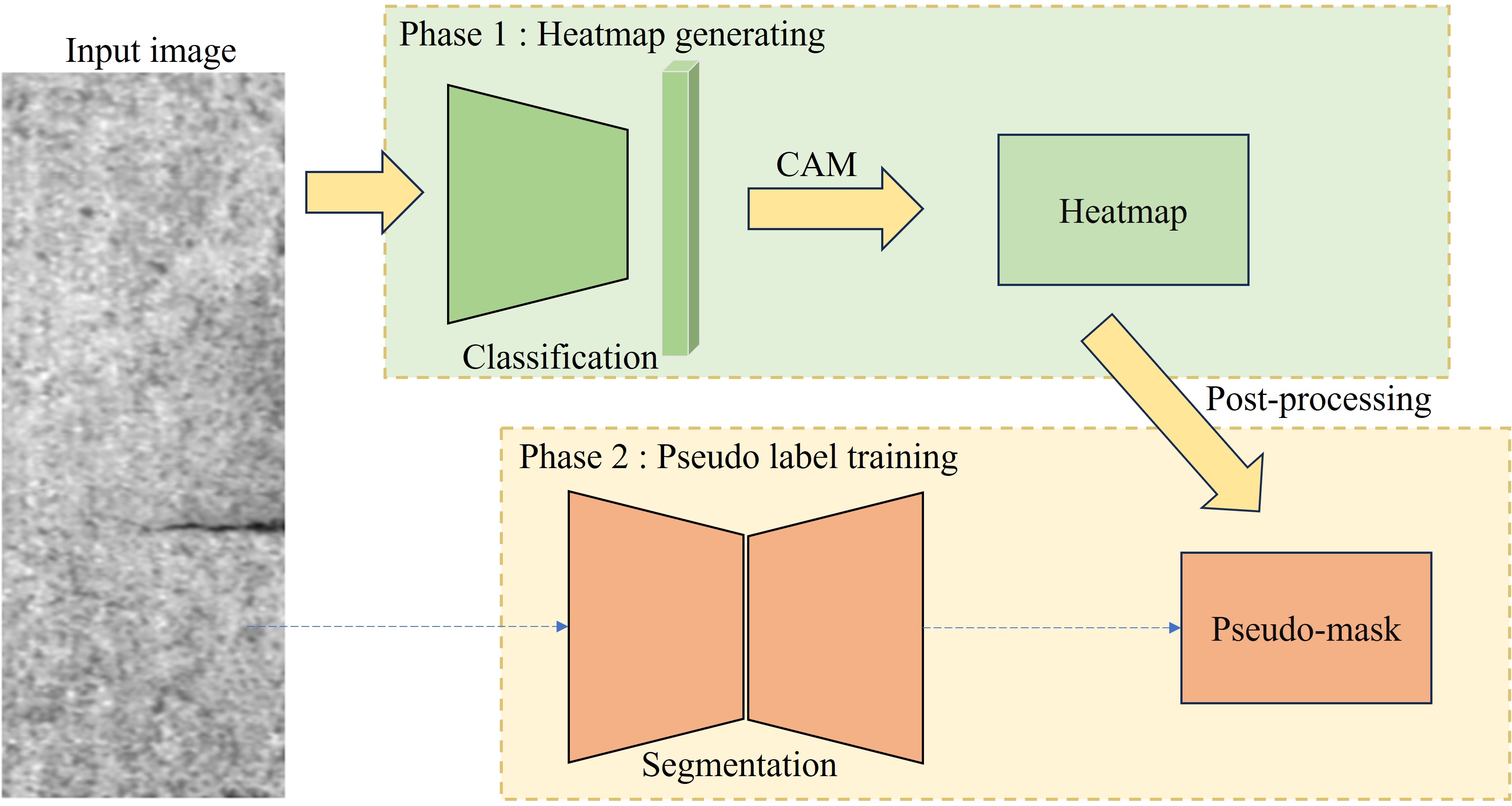}
    \caption{The workflow of weakly-supervised defect segmentation.}
    \label{fig:workflow}
\end{figure}

In this section, we present the proposed weakly supervised defect segmentation method. First, as shown in Fig.\ref{fig:workflow}, we divide the weakly supervised defect segmentation into two components, namely initial heatmap (also referred to as saliency map) generation followed by pseudo-label training. In the first phase, we propose RA-CAM based on filtering-guided backpropagation (FGBP) for generating high-quality and high-resolution heat maps. In the second phase, we utilize the heatmaps generated before, which are processed to generate pseudo-labels for training fully supervised models. We subsequently provide a stepwise explanation of FGBP, RA-CAM, and pseudo-label training.

\subsection{Filtering-Guided Backpropagation}

In this section, we introduce the filtering-guided backpropagation. (FGBP). Mathematically, let $f$ denote the CNN with $L$ convolutional layers, whose parameters are $\bm{\theta}$. For a given input image $\bm{I}$, the output is $y \in \mathbb R^c$ with category $c$, the prediction $y^c$ before the softmax can be obtained by
\begin{equation}
    y^c = f^c(\bm{I};\bm{\theta}). 
\end{equation}
Let $\bm{A}^{l}\in \mathbb{R}^{W_l \times H_l \times C_l}$ denote the feature map of the $k$-th channel generated by the $l$-th layer in the CNN, $l\in {1, 2, ..., L}$, where $W_l$ and $H_l$ are the width and height of $l$-th feature map respectively, and $C_l$ is the number of the channels in $l$-th convolutional layer. The gradient of output score $y^c$ with respect to the activation $\bm{A}^{kl}$ at location $(i,j)$ is $ g_{ij}^{ckl} = \frac{\partial y^c}{\partial A_{ij}^{kl}} $. 

For the forward propagation process of neural networks, we have

\begin{equation}
    A_l = relu(A_{l-1}) = max(A_{l-1},0). 
\end{equation}
Consequently, in the conventional process of backpropagation, we are able to derive

\begin{equation}
    R^l_i = (A_i > 0) \cdot R^{l+1}_i, 
\end{equation}
where $R^{l+1}_i = \frac{\partial y}{\partial A^{l+1}_i} $. Then the guided backpropagation is:

\begin{equation}
    R^l_i = (A_i > 0) \cdot (R^{l+1}_i > 0) \cdot R^{l+1}_i. 
\end{equation}

Moreover, through the specification of the hyperparameter $\delta$, we introduce an adaptive guided backpropagation process, which can be articulated as follows:

\begin{equation}
    R^l_i = (A_i > 0) \cdot (R^{l+1}_i > \delta) \cdot R^{l+1}_i. 
\end{equation}

\begin{figure}[ht]
    \centering
    \includegraphics[width=1\linewidth]{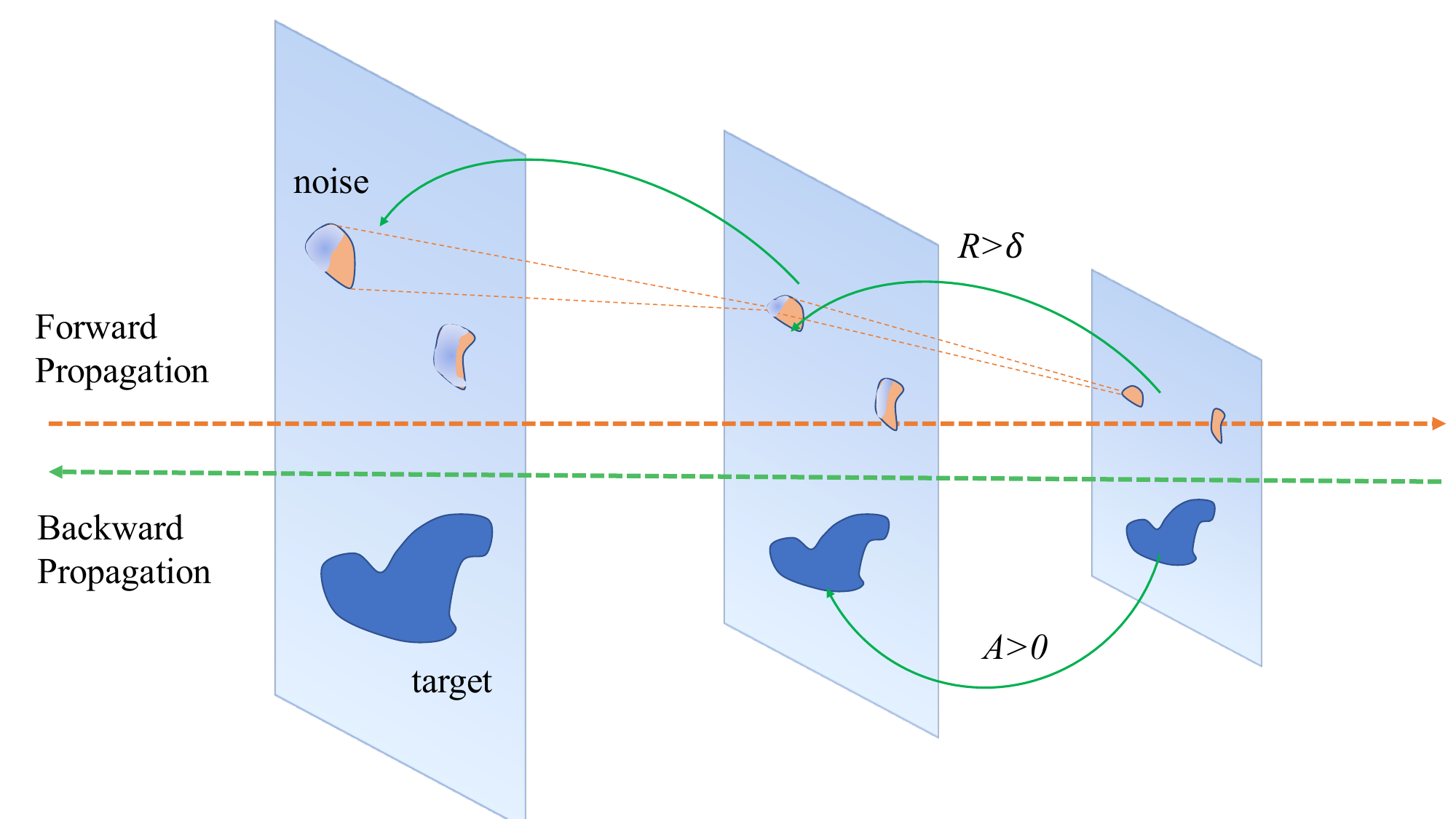}
    \caption{The workflow of weakly-supervised defect segmentation.}
    \label{fig:fgbp}
\end{figure}

As shown in Fig.\ref{fig:fgbp}, we present the computational workflow of FGBP. Since regions with lower gradient values are more likely to correspond to non-target noise areas, truncating low-value regions during the backpropagation process enables progressive removal of background noise.

\subsection{{Region-Aware Class
Activation Maps.}}
\textbf{1) Hierarchical semantic information.}
Traditionally, methods of the CAM variety take the output from the final convolutional layer as the foundational feature map, merging it through diverse weighting strategies. Including pooling layers in convolutional neural networks often leads to the generation of saliency maps with reduced resolution and a loss of detailed information. Illustrated in the Fig.\ref{fig1:gradcam_KSDD}, we have employed the GradCAM technique to produce saliency maps for the outputs of the five stages within the VGG16 network. Notably, from stage S1 through to S5, there is a progressive reduction in the resolution of the defect features. Concurrently, there is a consistent increase in feature intensity, accompanied by a reduction in spurious features in the background regions.

\textbf{2) RA-CAM.}Based on the aforementioned analysis, we propose the region-aware class activation map (RA-CAM). The specific computational process is shown in the figure. Formally, we have:

\begin{equation}
    R_{\text{CAM}}^{cl} = R_{\text{CAM}}^{cl}(\bm{x},\mathbb{I} (\frac{\partial y^c}{\partial \bm{A}_{l}} \geq \delta) \otimes \frac{\partial y^c}{\partial \bm{A}_{l}} ),
    \label{eq9}
\end{equation}
where $\mathbb{I}(\cdot)$ is the indicator function, $\bm{A}_{l}$ is the target feature map in the $l-th$ layer, and $\delta$ is a hyperparameter that is set to the $\delta$-th percentile of the positive values in each feature map. Eventually, we obtain the heatmap by

\begin{equation}
    M_{\text{RA-CAM}}^{cl} = \text{ReLU} (\sum_k R_{kl}^c \cdot \bm{A}^{kl}).
    \label{eq10}
\end{equation}

It is crucial to highlight that our method stands distinct from Guided Grad-CAM at a fundamental level, rather than being a simple augmentation with an adaptive module. Guided Grad-CAM \cite{guidedgradcam} merges the outcomes of guided gradient generation with Grad-CAM through element-wise multiplication. In contrast, RA-CAM introduces an innovative feature weighting strategy for Class Activation Maps (CAM), offering a fresh perspective in the realm of feature visualization and segmentation.

\subsection{{Pseudo Label Training}.} 

As shown in Fig.\ref{fig:workflow}, the comprehensive workflow of weakly supervised semantic segmentation is simplified into two main stages. The first stage involves heat map generation, and the second stage consists of pseudo-label training (generating pseudo-labels can also be considered as the second stage). In the first stage, we first train a classification model using image-level annotated data. Subsequently, we employ the novel RA-CAM method to generate heat maps. Finally, image processing techniques are applied to enhance defect regions and obtain pseudo-labels. In the second stage, we select high-performance, fully supervised semantic segmentation models and directly train them using the pseudo-labels. Additionally, details of post-processing are described below.




\underline{Post-processing}: Once the heatmaps have been generated, they can be further processed using image processing techniques to segment the regions of interest. In this study, we have adopted an adaptive thresholding approach that leverages Otsu's method for segmenting the highlighted areas.

\section{Experiments}

\subsection{Datasets Descriptions}
We evaluated the proposed weakly supervised semantic segmentation method on two widely used defect detection datasets. The specifications of these datasets are detailed as follows.

\textbf{1) KSDD dataset.}  The KolektorSDD dataset~\cite{ksdd}, developed through collaboration with the Kolektor Group, focuses on electrical commutator surface defects with expert annotations. Original image dimensions maintain a fixed width of 500 pixels while exhibiting height variations between 1240 and 1270 pixels. This benchmark contains 399 annotated samples, comprising 52 defective instances and 347 defect-free cases, systematically collected from 50 distinct physical commutator units. 


\textbf{2) KSDD2 dataset.}  The second dataset
, KSDD2 (Kolektor Surface Defect Detection Dataset Version 2)~\cite{ksdd2}, comprises 3,012 high-resolution industrial inspection images with nominal dimensions of 230×630 pixels. This industrial-grade dataset is partitioned into a training set containing 2,085 defect-free samples (246 defective cases) and a test set comprising 894 non-defective specimens (110 defective instances), maintaining an 8:2 split ratio between training and evaluation subsets.




\subsection{Experimental Configurations}

\textbf{1) Implementation settings.} 
In the classification model training phase, we implemented VGG16 as the backbone network with leaky ReLU activation functions. The model was optimized using stochastic gradient descent (SGD) with data augmentation through random horizontal and vertical flipping. All input images were resized to standardized dimensions: 512×1408 pixels for KSDD and 224×640 pixels for KSDD2, maintaining consistent aspect ratios through proportional scaling. We configured the batch size as 4 for the pseudo-label training of the segmentation network and initialized the learning rate at 0.0005. The SGD optimizer was employed with a momentum coefficient of 0.9 to accelerate gradient updates in relevant directions, thereby enhancing network convergence efficiency. The image resizing protocol remained identical to that of the classification stage.

All algorithms were implemented using PyTorch 1.12 and executed on a computational cluster equipped with an Intel(R) Xeon(R) Silver 4310 CPU and NVIDIA RTX A6000 GPUs. The experimental environment ensured CUDA 11.6 compatibility for hardware acceleration.



\textbf{2) Evaluation metrics.} For quantitative evaluation of segmentation performance, we adopt a comprehensive metric suite comprising Intersection over Union (IoU), mean IoU (mIoU), Precision, Recall, and F1-score. Specifically, the class-specific IoU for the defective category was reported to assess the proposed RA-CAM, while the subsequent pseudo-label segmentation quality was evaluated through both IoU and mIoU to holistically measure the integrated pipeline's efficacy. The evaluation protocol employed three complementary metrics: Precision (positive predictive value), Recall (true positive rate), and their harmonic mean (F1-score), providing multi-dimensional performance characterization. The metrics are formulated as follows:
\begin{equation}
\label{IoU}
\text{IoU} = \frac{TP}{TP + FP + FN}
\end{equation}

\begin{equation}
\text{mIoU} = \frac{1}{|\mathcal{C}|} \sum_{c \in \mathcal{C}} \frac{TP_c}{TP_c + FP_c + FN_c}
\end{equation}

\begin{equation}
\label{Precision}
\text{Precision} = \frac{TP}{TP + FP}
\end{equation}

\begin{equation}
\label{Recall}
\text{Recall} = \frac{TP}{TP + FN}
\end{equation}


\begin{equation}
\label{F1}
\text{F1 score} = \frac{2 \times \text{Precision} \times \text{Recall}}{\text{Precision} + \text{Recall}},
\end{equation}
where $\mathcal{C}$ is the set of semantic classes ($|\mathcal{C}|$ = total classes). $TP_c$ represent the true positives for class $c$, $FP_c$ are the false positives for class $c$, and $FN_c$ are the false negatives for class $c$.

\subsection{Main Results of Weakly-Supervised Segmentation}

Here, we compare the performance of the semantic segmentation results of the proposed RA-CAM with other sota methods. 

\textbf{1) Results on the KSDD dataset.}
The comparison of segmentation performance on the KSDD dataset is shown in Tab. \ref{tab1}. It can be seen that our proposed RA-CAM method surpasses the previously best method, Ablation-CAM, by 7.24\% in the IoU metric. It is particularly worth mentioning that RA-CAM has made progress of 8.95\% and 12.01\% over Layer-CAM and FullGrad methods, respectively. Both of these methods incorporate hierarchical semantic information, indicating that merely combining hierarchical information may not necessarily enhance segmentation performance and that further extraction of semantic information at different levels is still required. Therefore, this also strongly proves the effectiveness of the filtering-guided gradient proposed in this paper for the extraction of target semantic information.

\begin{figure}[ht]
    \centering
    \includegraphics[scale=0.2]{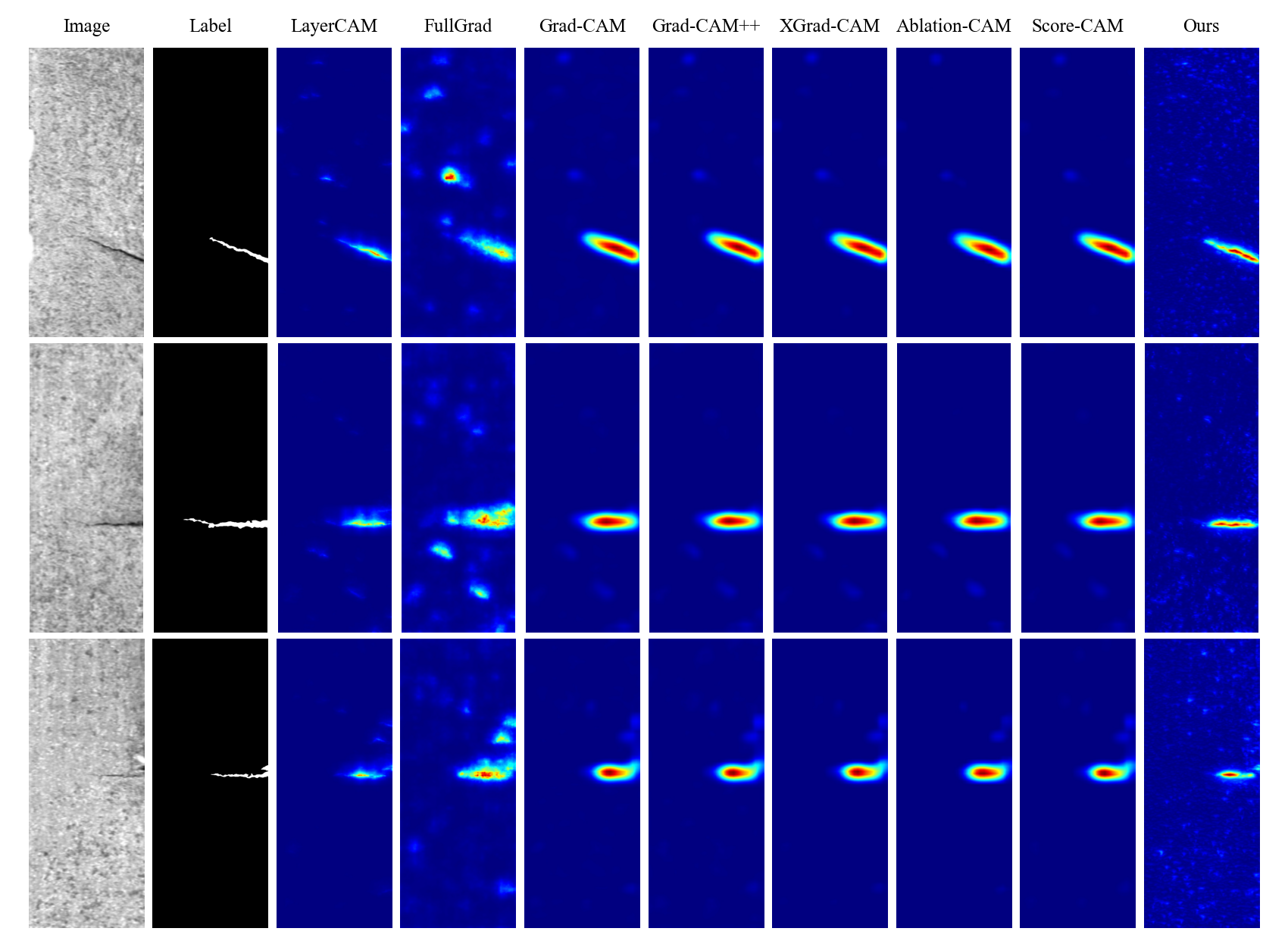}
    \caption{Comparison of heatmaps generated by different weakly supervised semantic segmentation methods on the KSDD dataset.}
    \label{fig:gradcam_KSDD}
\end{figure}

\begin{figure}[ht]
    \centering
    \includegraphics[scale=0.17]{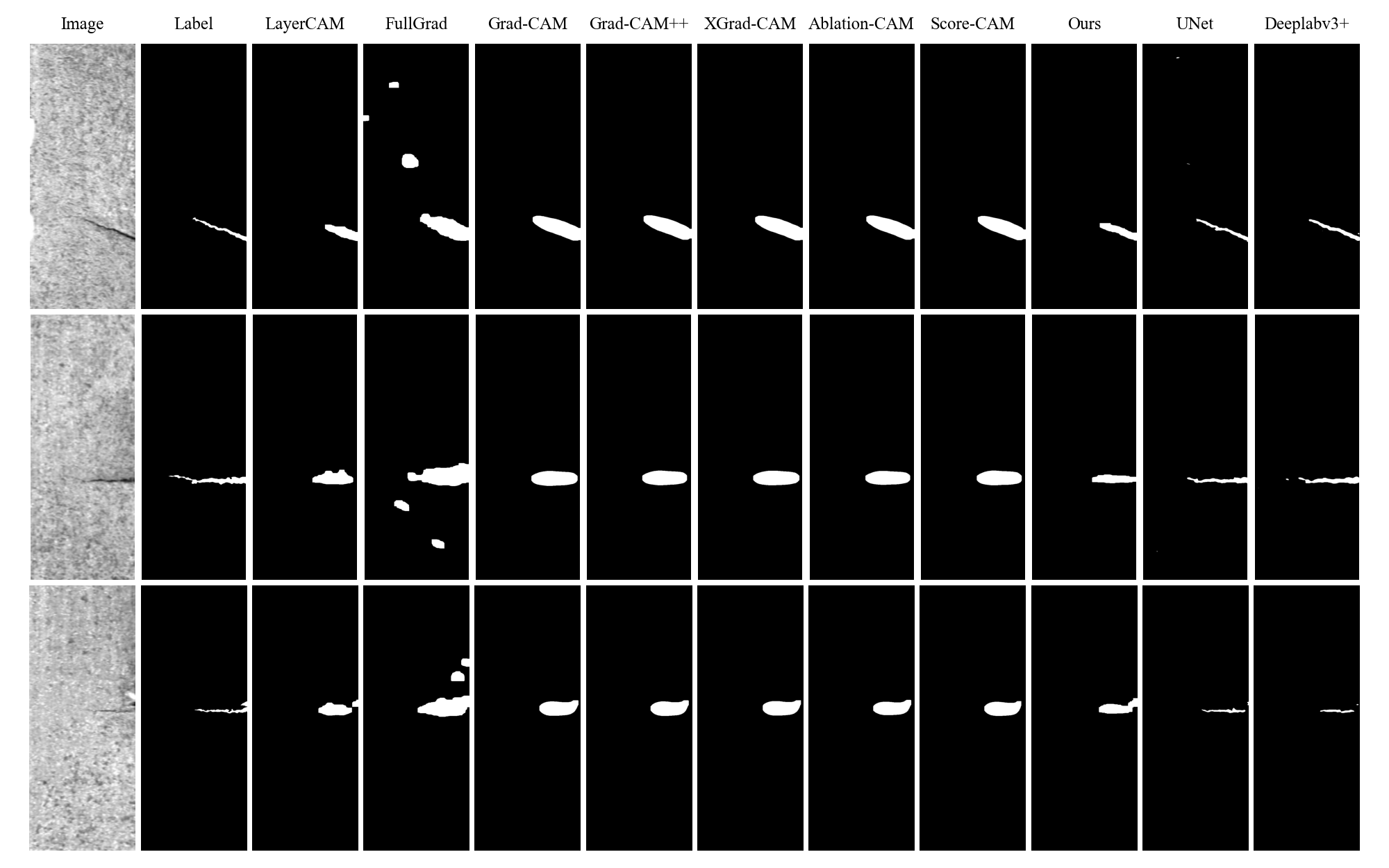}
    \caption{Comparison of segmentation results generated by different weakly supervised semantic segmentation methods on the KSDD dataset.}
    \label{KSDD_seg}
\end{figure}

Fig.\ref{fig:gradcam_KSDD} demonstrates the saliency maps generated by different interpretation methods. Critical observation reveals that approaches lacking hierarchical semantic integration (e.g., Grad-CAM) produce coarse activation patterns with insufficient detail resolution, failing to preserve high-frequency spatial information. In contrast, methods incorporating multi-level semantic features (e.g., LayerCAM and FullGrad) generate refined edge delineation, yet remain susceptible to background noise contamination. Specifically, LayerCAM exhibits constrained feature localization with under-activated regions, whereas FullGrad demonstrates excessive spurious activations due to incomplete feature weighting regularization. As shown in Fig.\ref{KSDD_seg}, while traditional fully-supervised approaches achieve better boundary accuracy through dense pixel-level supervision, our approach obtains higher-resolution defect segmentation performance using only categorization-level labels and also provides high-quality pseudo-labels for downstream learning tasks. This label efficiency highlights the effectiveness of our approach.

\begin{table}
 \centering
\caption{
The weakly supervised semantic segmentation performance on the KSDD dataset. The results in bold indicate the best performance.
}\label{tab1}
\begin{tabular}{c|c|c|c|c}
\hline
method & IoU(\%) & Precision(\%) & Recall(\%) & Micro-F1(\%)\\
\hline
Grad-CAM &17.50&	19.27&	65.56&	29.79\\
Grad-CAM++ &17.22&	18.97&	65.14&	29.38\\
XGrad-CAM &17.06& 	18.86& 	64.10& 	29.15 \\
Ablation-CAM &17.96& 	19.97& 	64.08& 	30.46 \\
Score-CAM &17.63&	19.50&	64.76&	29.98\\
Layer-CAM &16.25& 	17.76& 	65.61& 	27.95 \\
FullGrad &13.19&	13.58&	{\bfseries 82.21}&	23.31\\
RA-CAM &{\bfseries 25.20}&	{\bfseries 30.74}&	58.29&	{\bfseries 40.25}\\

\hline
\end{tabular}
\end{table}

\begin{figure}[ht]
    \centering
    \includegraphics[scale=0.2]{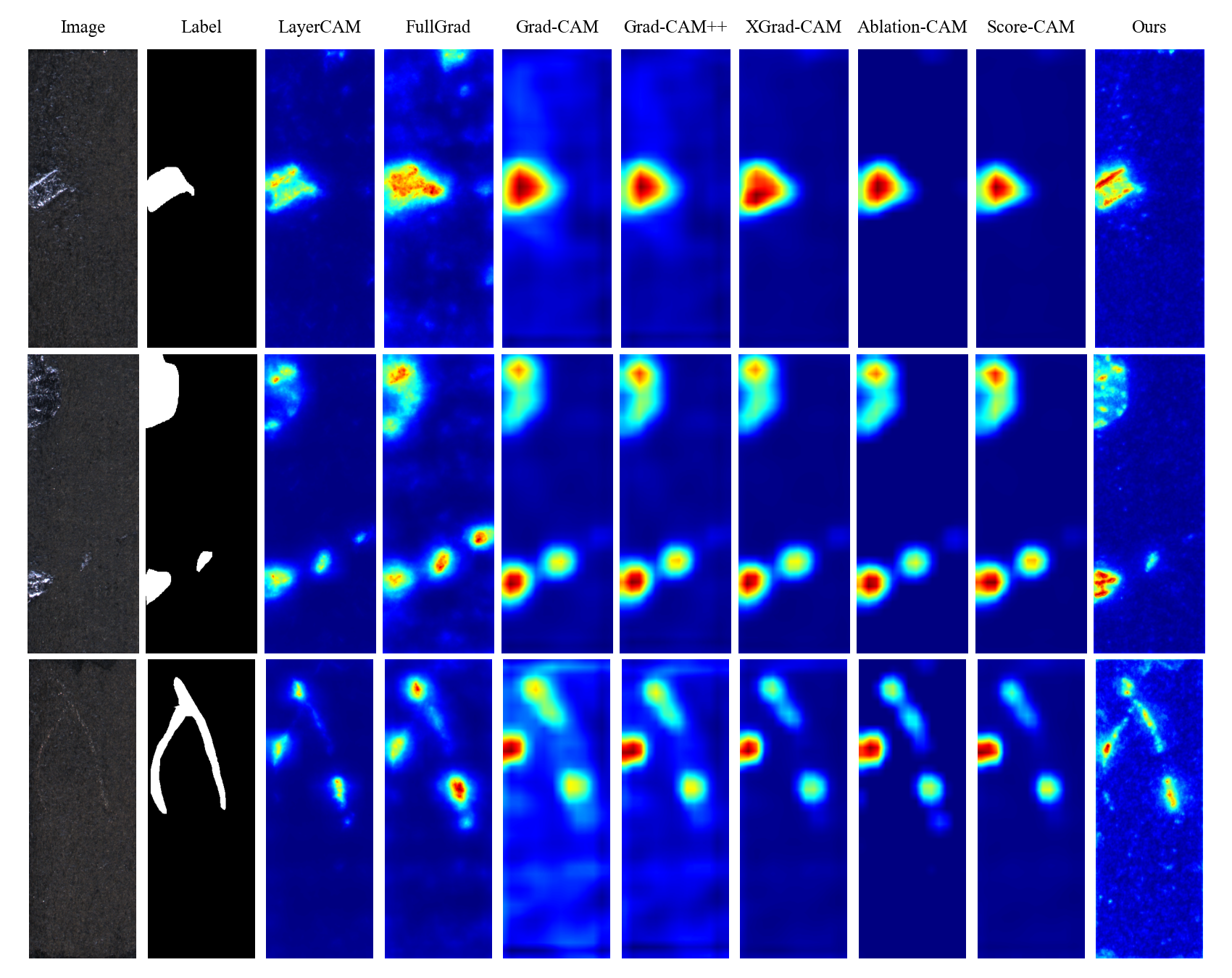}
    \caption{Comparison of heatmaps generated by different weakly supervised semantic segmentation methods on the KSDD2 dataset.}
    \label{fig:gradcam_KSDD2}
\end{figure}

\textbf{2) Results on the KSDD2 dataset.} On the KSDD2 dataset, the segmentation performance of the state-of-the-art (SOTA) methods and our method is shown in Tab.\ref{tab2}. The segmentation algorithm we proposed achieved an IoU of 45.54\%, surpassing the previous SOTA method LayerCAM by 3.53\%, and exceeding FullGrad by 7.59\%. Additionally, our method also achieved an F1 score of 62.58\%, surpassing LayerCAM by 3.41\% and ScoreCAM by 4.62\%. As shown in Fig.\ref{fig:gradcam_KSDD2}, we present some saliency maps generated by several weakly supervised methods, where the highlighted areas indicate the location of the target. It can be observed that our proposed method not only provides more complete highlighted areas but also offers higher resolution.

\begin{table}
 \centering
\caption{
The segmentation performance on the KSDD2 dataset. The results in bold indicate the best performance.
}\label{tab2}
\begin{tabular}{c|c|c|c|c}
\hline
method & IoU(\%) & Precision(\%) & Recall(\%) & Micro-F1(\%)\\
\hline
Grad-CAM & 38.67 & 49.19 & 64.41 & 55.78\\
Grad-CAM++ &35.17	& 44.27&	63.10 & 52.04\\
XGrad-CAM &  39.67&	51.03&	64.06&	56.81\\
Ablation-CAM &  26.06 &	31.02 &	61.99 &	41.35\\
Score-CAM & 40.80&	56.72&	59.25&	57.96\\
Layer-CAM & 42.01&	60.46&	57.92&	59.17\\
FullGrad &37.95&	46.29&	{\bfseries 67.81}&	55.02\\
RA-CAM & {\bfseries 45.54}&	{\bfseries 71.18}&	55.84&	{\bfseries 62.58}\\

\hline
\end{tabular}
\end{table}

\begin{figure}[ht]
    \centering
    \includegraphics[scale=0.17]{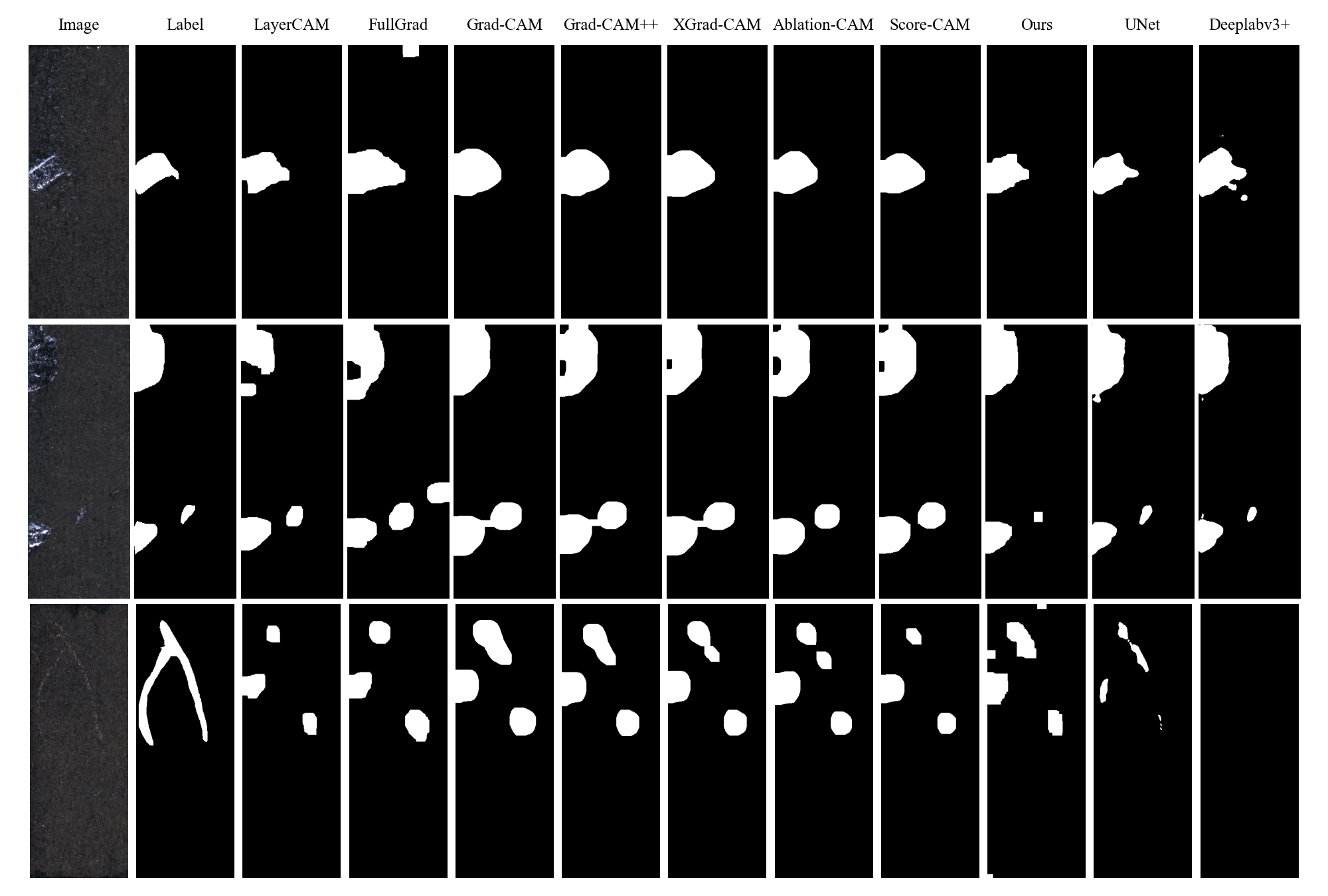}
    \caption{Comparison of segmentation results generated by different weakly supervised semantic segmentation methods on the KSDD2 dataset.}
    \label{SDD2_seg}
\end{figure}

Consistently, Fig.\ref{fig:gradcam_KSDD2} and Fig.\ref{SDD2_seg} comparatively visualize the class activation maps and final segmentation outputs across different weakly supervised approaches. Beyond conclusions consistent with KSDD observations, our analysis reveals fundamental limitations in fully supervised semantic segmentation. As evidenced in Fig.\ref{SDD2_seg}, DeeplabV3+ exhibits systematic under-segmentation errors even under full supervision, a fundamental limitation rooted in its loss function design. The conventional cross-entropy loss in semantic segmentation frameworks imposes a homogeneous penalty distribution across defect pixels, creating fundamental limitations in defect detection scenarios. While minor segmentation inaccuracies on individual pixels yield negligible loss increments, these localized errors can critically manifest as false negatives in defect-free specimen classification. More critically, the inherent class imbalance, where defect pixels often constitute less than 5\% of the total image area, induces systematic model bias towards the majority of non-defective regions, significantly attenuating sensitivity to subtle defect patterns. It also reflects the unique advantages of weakly supervised segmentation in defect detection tasks.

\subsection{Main Results of Pseudo-Label Training}

This section presents the principal findings of the pseudo-label training process, as summarized in Tab.\ref{tab:psedoseg}. The experimental results demonstrate substantial performance improvement through pseudo-label training, with IoU scores reaching 37.86\% and 57.56\% on the KSDD and KSDD2 datasets, respectively. Compared with fully supervised baselines, our weakly supervised method achieves 88.6\% of DeepLabV3+’s mIoU performance on KSDD, while approaching the performance level of fully supervised models on KSDD2. These empirical findings confirm that pseudo-label training serves as a pivotal mechanism for performance enhancement in weakly supervised semantic segmentation frameworks.

\begin{table}[ht]
    \centering
    \begin{tabular}{c|c|c cc c}
    \hline
      \multirow{2}{*}{method}   &\multirow{2}{*}{type} &\multicolumn{2}{c}{KolektorSDD(\%)}& \multicolumn{2}{c}{KolektorSDD2(\%)} \\
    
               &     &  Defect   &        mIoU          & Defect & mIoU\\
    \hline
        UNet   & FS & 54.56  &  77.28 &    58.79& 79.40  \\
        deeplabv3 &  FS&  53.89  &  76.95 &    59.98& 79.99  \\
        deeplabv3+& FS &  55.59  &  77.80 &  61.08& 80.54  \\
        RACAM+deeplabv3(Ours)& WS &  37.86  & 68.93  &  57.56& 78.78 \\
    \hline
    \end{tabular}
    \caption{Comparison of segmentation results between fully supervised models and our proposed pseudo-label training}
    \label{tab:psedoseg}
\end{table}


\subsection{Ablation Study}
\begin{figure}[ht]
    \centering
    \includegraphics[scale=0.5]{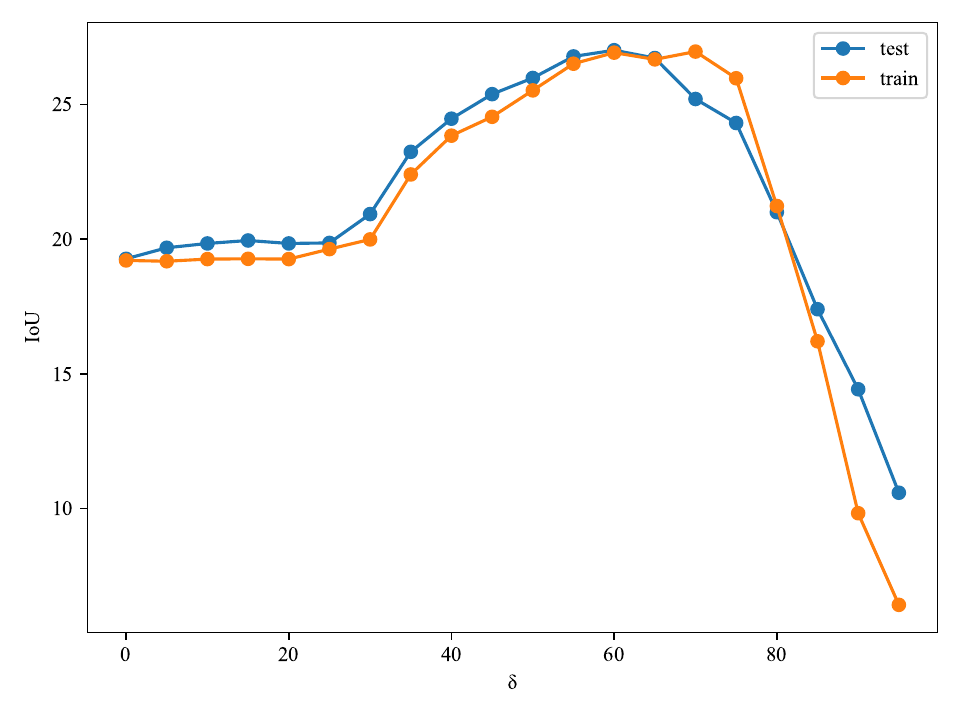}
    \caption{The trend of segmentation performance on the KSDD dataset with the variation of hyperparameter $\delta$.}
    \label{fig:hyperKSDD}
\end{figure}

\begin{figure}[ht]
    \centering
    \includegraphics[scale=0.5]{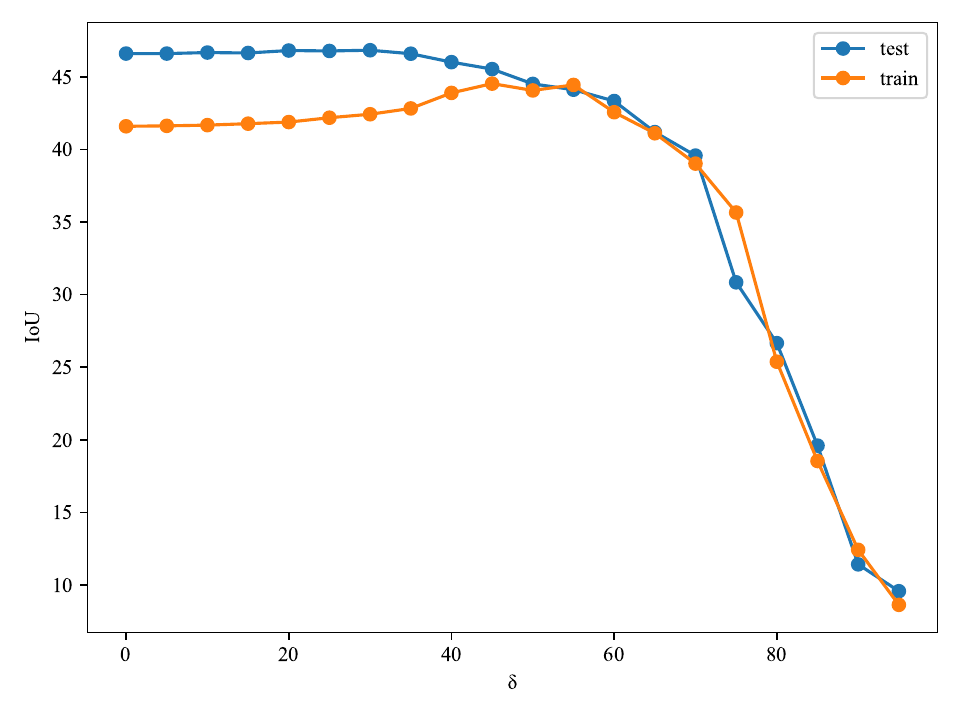}
    \caption{The trend of segmentation performance on the KSDD2 dataset with the variation of hyperparameter $\delta$.}
    \label{fig:hyperKSDD2}
\end{figure}

\textbf{1) Effect of the hyperparameter in filtering-guide backpropagation.}
As shown in Fig.\ref{fig:hyperKSDD} and Fig.\ref{fig:hyperKSDD2}, we have plotted the changes in IoU on both the training and testing sets of the KSDD and KSDD2 datasets, respectively, with varying values of $\delta$. It can be observed that on both datasets, the curve trends on the training and testing sets are consistent, allowing us to select an appropriate threshold based on the training set. Another piece of experience is that the RA-CAM performs better around the $\delta$ value of 50\%, so this value can be taken as the default threshold. Upon further analysis, it can be observed that in Fig.\ref{fig:hyperKSDD}, the performance of IoU first increases and then decreases as $\delta$ increases. A similar trend is observed in Fig.\ref{fig:hyperKSDD2}, although the increase is less pronounced. This is because as the $\delta$ increases, the background area is gradually removed, and the target area becomes more prominent. However, after reaching a certain threshold, the target area also begins to be removed. This suggests that there is an optimal range for $\delta$ that balances the removal of background noise without compromising the integrity of the target area, which is crucial for peak performance in segmentation tasks. 



\textbf{2) Using filtering-guide backpropagation as a plug-in.}

The proposed filtering-guided backpropagation (FGBP) is implemented as a plug-and-play module that effectively supplants conventional gradient backpropagation approaches in analogous architectures. As shown in Tab.\ref{tab3}, experimental results indicate that integrating FGBP with established methods (FullGrad and LayerCAM) yields consistent performance improvements across both KSDD and KSDD2 datasets. Notably, optimal threshold selection remains crucial for FGBP implementation, with specific parameter configuration guidelines detailed in the above ablation study. This empirical evidence substantiates the considerable potential of FGBP to serve as a performance-enhancing component for gradient-based weakly supervised methodologies. 

\begin{table}[h]
    \centering
    \begin{tabular}{ccc}
    \hline
              & KolektorSDD(IoU,\%) & KolektorSDD2(IoU,\%) \\
    \hline
     FullGrad & 13.19               & 37.95        \\ 
     FullGrad+FGBP & {\bfseries 16.73}     & {\bfseries 41.76} \\
     LayerCAM & 16.25               & 42.01        \\
     LayerCAM+FGBP & {\bfseries18.65} & {\bfseries 42.95}        \\
     \hline
    \end{tabular}
    \caption{Results for the ablation experiments of LayerCAM and FullGrad.}
    \label{tab3}
\end{table}


\section{Conclusion}

In this work, we have proposed region-aware class activation maps for weakly supervised defect segmentation tasks. In order to obtain more accurate weights for the target region and reduce the influence of background and noise, we design filter-guided backpropagation. Furthermore, the region-aware weighting is proposed. Experimental results show that our proposed RA-CAM can extract finer target regions and the designed backpropagation method can be applied to other similar methods such as LayerCAM. It is worth mentioning that we also analyze the unfitness of semantic segmentation algorithms for defect detection tasks, demonstrating the superiority of weakly supervised approaches.

\bibliographystyle{IEEEtran}
\bibliography{bio.bib}

\begin{thebibliography}{10}
\providecommand{\url}[1]{#1}
\csname url@samestyle\endcsname
\providecommand{\newblock}{\relax}
\providecommand{\bibinfo}[2]{#2}
\providecommand{\BIBentrySTDinterwordspacing}{\spaceskip=0pt\relax}
\providecommand{\BIBentryALTinterwordstretchfactor}{4}
\providecommand{\BIBentryALTinterwordspacing}{\spaceskip=\fontdimen2\font plus
\BIBentryALTinterwordstretchfactor\fontdimen3\font minus \fontdimen4\font\relax}
\providecommand{\BIBforeignlanguage}[2]{{%
\expandafter\ifx\csname l@#1\endcsname\relax
\typeout{** WARNING: IEEEtran.bst: No hyphenation pattern has been}%
\typeout{** loaded for the language `#1'. Using the pattern for}%
\typeout{** the default language instead.}%
\else
\language=\csname l@#1\endcsname
\fi
#2}}
\providecommand{\BIBdecl}{\relax}
\BIBdecl

\bibitem{gao2022review}
Y.~Gao, X.~Li, X.~V. Wang, L.~Wang, and L.~Gao, ``A review on recent advances in vision-based defect recognition towards industrial intelligence,'' \emph{Journal of Manufacturing Systems}, vol.~62, pp. 753--766, 2022.

\bibitem{GOLNABI2007630}
\BIBentryALTinterwordspacing
H.~Golnabi and A.~Asadpour, ``Design and application of industrial machine vision systems,'' \emph{Robotics and Computer-Integrated Manufacturing}, vol.~23, no.~6, pp. 630--637, 2007, 16th International Conference on Flexible Automation and Intelligent Manufacturing. [Online]. Available: \url{https://www.sciencedirect.com/science/article/pii/S0736584507000233}
\BIBentrySTDinterwordspacing

\bibitem{ren2022state}
Z.~Ren, F.~Fang, N.~Yan, and Y.~Wu, ``State of the art in defect detection based on machine vision,'' \emph{International Journal of Precision Engineering and Manufacturing-Green Technology}, vol.~9, no.~2, pp. 661--691, 2022.

\bibitem{ksdd2}
J.~Bozic, D.~Tabernik, and D.~Skocaj, ``Mixed supervision for surface-defect detection: From weakly to fully supervised learning,'' \emph{Comput. Ind.}, vol. 129, p. 103459, 2021.

\bibitem{longtail}
S.~Zhang, Z.~Li, S.~Yan, X.~He, and J.~Sun, ``Distribution alignment: A unified framework for long-tail visual recognition,'' in \emph{Proceedings of the IEEE/CVF conference on computer vision and pattern recognition}, 2021, pp. 2361--2370.

\bibitem{od2022review}
H.~M. Ahmad and A.~Rahimi, ``Deep learning methods for object detection in smart manufacturing: A survey,'' \emph{Journal of Manufacturing Systems}, vol.~64, pp. 181--196, 2022.

\bibitem{sultana2020evolution}
F.~Sultana, A.~Sufian, and P.~Dutta, ``Evolution of image segmentation using deep convolutional neural network: A survey,'' \emph{Knowledge-Based Systems}, vol. 201, p. 106062, 2020.

\bibitem{liu2023survey}
Y.~Liu, C.~Zhang, and X.~Dong, ``A survey of real-time surface defect inspection methods based on deep learning,'' \emph{Artificial Intelligence Review}, vol.~56, no.~10, pp. 12\,131--12\,170, 2023.

\bibitem{tao2022deep}
X.~Tao, X.~Gong, X.~Zhang, S.~Yan, and C.~Adak, ``Deep learning for unsupervised anomaly localization in industrial images: A survey,'' \emph{IEEE Transactions on Instrumentation and Measurement}, vol.~71, pp. 1--21, 2022.

\bibitem{CAM}
B.~Zhou, A.~Khosla, {\`{A}}.~Lapedriza, A.~Oliva, and A.~Torralba, ``Learning deep features for discriminative localization,'' in \emph{2016 {IEEE} Conference on Computer Vision and Pattern Recognition, {CVPR} 2016, Las Vegas, NV, USA, June 27-30}, 2016, pp. 2921--2929.

\bibitem{CADN}
J.~Zhang, H.~Su, W.~Zou, X.~Gong, Z.~Zhang, and F.~Shen, ``Cadn: A weakly supervised learning-based category-aware object detection network for surface defect detection,'' \emph{Pattern Recognition}, vol. 109, p. 107571, 2021.

\bibitem{GradCAM}
R.~R. Selvaraju, M.~Cogswell, A.~Das, R.~Vedantam, D.~Parikh, and D.~Batra, ``Grad-cam: Visual explanations from deep networks via gradient-based localization,'' in \emph{{IEEE} International Conference on Computer Vision, {ICCV} 2017, Venice, Italy, October 22-29}, 2017, pp. 618--626.

\bibitem{layercam}
P.~Jiang, C.~Zhang, Q.~Hou, M.~Cheng, and Y.~Wei, ``Layercam: Exploring hierarchical class activation maps for localization,'' \emph{{IEEE} Trans. Image Process.}, vol.~30, pp. 5875--5888, 2021.

\bibitem{fan}
F.-L. Fan, J.~Xiong, M.~Li, and G.~Wang, ``On interpretability of artificial neural networks: A survey,'' \emph{IEEE Transactions on Radiation and Plasma Medical Sciences}, vol.~5, no.~6, pp. 741--760, 2021.

\bibitem{Grad-cam++}
A.~Chattopadhay, A.~Sarkar, P.~Howlader, and V.~N. Balasubramanian, ``Grad-cam++: Generalized gradient-based visual explanations for deep convolutional networks,'' in \emph{2018 IEEE winter conference on applications of computer vision (WACV)}.\hskip 1em plus 0.5em minus 0.4em\relax IEEE, 2018, pp. 839--847.

\bibitem{Axiom-based}
R.~Fu, Q.~Hu, X.~Dong, Y.~Guo, Y.~Gao, and B.~Li, ``Axiom-based grad-cam: Towards accurate visualization and explanation of cnns,'' in \emph{31st British Machine Vision Conference 2020, {BMVC} 2020, Virtual Event, UK, September 7-10}, 2020.

\bibitem{liftcam}
H.~Jung and Y.~Oh, ``Towards better explanations of class activation mapping,'' in \emph{Proceedings of the IEEE/CVF International Conference on Computer Vision}, 2021, pp. 1336--1344.

\bibitem{RelevanceCAM}
J.~R. Lee, S.~Kim, I.~Park, T.~Eo, and D.~Hwang, ``Relevance-cam: Your model already knows where to look,'' in \emph{Proceedings of the IEEE/CVF Conference on Computer Vision and Pattern Recognition}, 2021, pp. 14\,944--14\,953.

\bibitem{LRP}
S.~Bach, A.~Binder, G.~Montavon, F.~Klauschen, K.-R. M{\"u}ller, and W.~Samek, ``On pixel-wise explanations for non-linear classifier decisions by layer-wise relevance propagation,'' \emph{PloS one}, vol.~10, no.~7, p. e0130140, 2015.

\bibitem{scorecam}
H.~Wang, Z.~Wang, M.~Du, F.~Yang, Z.~Zhang, S.~Ding, P.~Mardziel, and X.~Hu, ``Score-cam: Score-weighted visual explanations for convolutional neural networks,'' in \emph{2020 {IEEE/CVF} Conference on Computer Vision and Pattern Recognition, {CVPR} Workshops 2020, Seattle, WA, USA, June 14-19}, 2020, pp. 111--119.

\bibitem{ablationcam}
H.~G. Ramaswamy \emph{et~al.}, ``Ablation-cam: Visual explanations for deep convolutional network via gradient-free localization,'' in \emph{Proceedings of the IEEE/CVF Winter Conference on Applications of Computer Vision}, 2020, pp. 983--991.

\bibitem{Group-CAM}
Q.~Zhang, L.~Rao, and Y.~Yang, ``Group-cam: Group score-weighted visual explanations for deep convolutional networks,'' \emph{arXiv preprint arXiv:2103.13859}, 2021.

\bibitem{FSGCAM}
\BIBentryALTinterwordspacing
D.~Wang, Y.~Xia, W.~Pedrycz, Z.~Li, and Z.~Yu, ``Feature similarity group-class activation mapping (fsg-cam): Clarity in deep learning models and enhancement of visual explanations,'' \emph{Expert Systems with Applications}, p. 127553, 2025. [Online]. Available: \url{https://www.sciencedirect.com/science/article/pii/S0957417425011753}
\BIBentrySTDinterwordspacing

\bibitem{fullgrad}
S.~Srinivas and F.~Fleuret, ``Full-gradient representation for neural network visualization,'' in \emph{Advances in Neural Information Processing Systems 32: Annual Conference on Neural Information Processing Systems 2019, NeurIPS 2019, December 8-14, 2019, Vancouver, BC, Canada}, H.~M. Wallach, H.~Larochelle, A.~Beygelzimer, F.~d'Alch{\'{e}}{-}Buc, E.~B. Fox, and R.~Garnett, Eds., 2019, pp. 4126--4135.

\bibitem{zhou2025non}
X.~Zhou, Y.~Li, G.~Cao, and W.~Cao, ``Non-target feature filtering for weakly supervised semantic segmentation,'' \emph{Complex \& Intelligent Systems}, vol.~11, no.~1, pp. 1--15, 2025.

\bibitem{tursun2022sess}
O.~Tursun, S.~Denman, S.~Sridharan, and C.~Fookes, ``Sess: Saliency enhancing with scaling and sliding,'' in \emph{European Conference on Computer Vision}.\hskip 1em plus 0.5em minus 0.4em\relax Springer, 2022, pp. 318--333.

\bibitem{yolov11}
R.~Khanam and M.~Hussain, ``Yolov11: An overview of the key architectural enhancements,'' \emph{arXiv preprint arXiv:2410.17725}, 2024.

\bibitem{ren2016faster}
S.~Ren, K.~He, R.~Girshick, and J.~Sun, ``Faster r-cnn: Towards real-time object detection with region proposal networks,'' \emph{IEEE transactions on pattern analysis and machine intelligence}, vol.~39, no.~6, pp. 1137--1149, 2016.

\bibitem{UNet}
O.~Ronneberger, P.~Fischer, and T.~Brox, ``U-net: Convolutional networks for biomedical image segmentation,'' in \emph{International Conference on Medical Image Computing and Computer-Assisted Intervention}, 2015.

\bibitem{deeplabv3+}
L.~C. Chen, Y.~Zhu, G.~Papandreou, F.~Schroff, and H.~Adam, ``Encoder-decoder with atrous separable convolution for semantic image segmentation,'' in \emph{European Conference on Computer Vision}, 2018.

\bibitem{kaikouxiao}
J.~Zhong, Z.~Liu, Z.~Han, Y.~Han, and W.~Zhang, ``A cnn-based defect inspection method for catenary split pins in high-speed railway,'' \emph{IEEE Transactions on Instrumentation and Measurement}, vol.~68, no.~8, pp. 2849--2860, 2018.

\bibitem{cui2021sddnet}
L.~Cui, X.~Jiang, M.~Xu, W.~Li, P.~Lv, and B.~Zhou, ``Sddnet: A fast and accurate network for surface defect detection,'' \emph{IEEE Transactions on Instrumentation and Measurement}, vol.~70, pp. 1--13, 2021.

\bibitem{ma2024hierarchical}
J.~Ma, S.~Hu, J.~Fu, and G.~Chen, ``A hierarchical attention detector for bearing surface defect detection,'' \emph{Expert Systems with Applications}, vol. 239, p. 122365, 2024.

\bibitem{dong2019pga119}
H.~Dong, K.~Song, Y.~He, J.~Xu, Y.~Yan, and Q.~Meng, ``Pga-net: Pyramid feature fusion and global context attention network for automated surface defect detection,'' \emph{IEEE Transactions on Industrial Informatics}, vol.~16, no.~12, pp. 7448--7458, 2019.

\bibitem{yang2021automatic115}
L.~Yang, S.~Song, J.~Fan, B.~Huo, E.~Li, and Y.~Liu, ``An automatic deep segmentation network for pixel-level welding defect detection,'' \emph{IEEE Transactions on Instrumentation and Measurement}, vol.~71, pp. 1--10, 2021.

\bibitem{LIU2021102008fabric}
\BIBentryALTinterwordspacing
Z.~Liu, Z.~Huo, C.~Li, Y.~Dong, and B.~Li, ``Dlse-net: A robust weakly supervised network for fabric defect detection,'' \emph{Displays}, vol.~68, p. 102008, 2021. [Online]. Available: \url{https://www.sciencedirect.com/science/article/pii/S0141938221000226}
\BIBentrySTDinterwordspacing

\bibitem{liu2022explainable}
T.~Liu, H.~Zheng, J.~Bao, P.~Zheng, J.~Wang, C.~Yang, and J.~Gu, ``An explainable laser welding defect recognition method based on multi-scale class activation mapping,'' \emph{IEEE Transactions on Instrumentation and Measurement}, vol.~71, pp. 1--12, 2022.

\bibitem{guidedgradcam}
R.~R. Selvaraju, M.~Cogswell, A.~Das, R.~Vedantam, D.~Parikh, and D.~Batra, ``Grad-cam: visual explanations from deep networks via gradient-based localization,'' \emph{International journal of computer vision}, vol. 128, pp. 336--359, 2020.

\bibitem{ksdd}
D.~Tabernik, S.~Sela, J.~Skvarc, and D.~Skocaj, ``Segmentation-based deep-learning approach for surface-defect detection,'' \emph{Journal of Intelligent Manufacturing}, vol.~31, no.~3, pp. 759--776, 2020.

\end{thebibliography}

\vfill

\end{document}